\documentclass[times, twoside]{arxiv}
\usepackage{graphicx}
\graphicspath{ {./img/} }
\usepackage{xspace}
\usepackage{tabularx}
\usepackage{booktabs}
\usepackage{enumitem}
\setlist{nosep}

\usepackage{multirow}
\usepackage{array}
\usepackage{xr}

\newcolumntype{P}[1]{>{\raggedright\arraybackslash}p{#1}}

\leadauthor{Li}

\begin{document}

\title{Learning from Polar Representation: An Extreme-Adaptive Model for Long-Term Time Series Forecasting}
\shorttitle{Extreme-Adaptive Time Series Forecasting}

\author[1]{Yanhong Li}
\author[2]{Jack Xu}
\author[1]{David C. Anastasiu\thanks{Corresponding author: David C. Anastasiu. E-mail: \url{danastasiu@scu.edu}}}
\affil[1]{Department of Computer Science and Engineering, Santa Clara University, Santa Clara, CA, USA}
\affil[2]{Santa Clara Valley Water District, San Jose, CA, USA}

\maketitle
\begin{abstract}
In the hydrology field, time series forecasting is crucial for efficient water resource management, improving flood and drought control and increasing the safety and quality of life for the general population. However, predicting long-term streamflow is a complex task due to the presence of extreme events. It requires the capture of long-range dependencies and the modeling of rare but important extreme values. Existing approaches often struggle to tackle these dual challenges simultaneously. In this paper, we specifically delve into these issues and propose Distance-weighted Auto-regularized Neural network (DAN), a novel extreme-adaptive model for long-range forecasting of stremflow enhanced by polar representation learning. DAN utilizes a distance-weighted multi-loss mechanism and stackable blocks to dynamically refine indicator sequences from exogenous data, while also being able to handle uni-variate time-series by employing Gaussian Mixture probability modeling to improve robustness to severe events. We also introduce Kruskal-Wallis sampling and gate control vectors to handle imbalanced extreme data. On four real-life hydrologic streamflow datasets, we demonstrate that DAN significantly outperforms both state-of-the-art hydrologic time series prediction methods and general methods designed for long-term time series prediction.
\end{abstract}

\begin{figure*}[htb]
\centering
\includegraphics[width=\linewidth]{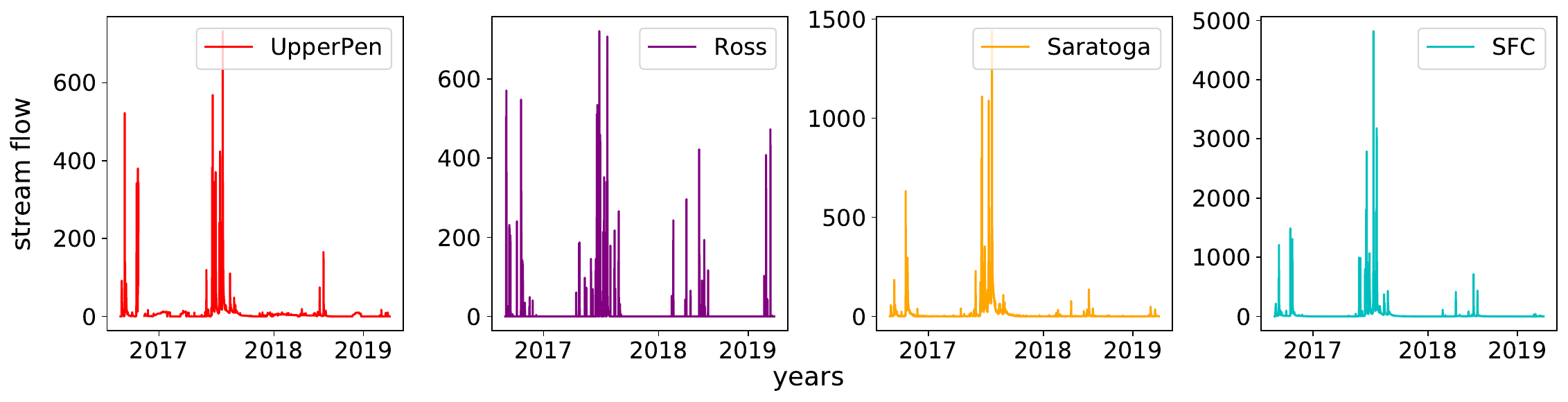}
\caption{Streamflow over 3 years.}\label{fig:stream_flow}
\end{figure*}
\section*{Introduction}

Time series forecasting has a critical role in diverse domains, enabling effective resource management and informed policy decisions. However, certain time series data pose a unique problem because they contain sporadic but significant extreme events, such as unexpected flash floods or climate change-induced droughts in the problem of streamflow prediction. The ability to forecast time series that include these types of extreme occurrences is an important research direction which has seen much attention in recent years~\cite{nguyen2004multiple,ding2019modeling,qi2020using,chen2020evanet,Yifan2021,li2022extremeadaptive}.

Traditionally, machine learning and statistics-based models were the basic foundation for time series prediction~\cite{box1970distribution,nielsen2019practical}. However, techniques like Autoregressive Integrated Moving Average (ARIMA)~\cite{boxjen76} seem to perform badly when dealing with large variations in the streamflow values, while other methods~\cite{shortridge2016machine,papacharalampous2022review,cheng2020long} are generally designed for short future horizon forecasting.

A variety of neural network architectures have been investigated for hydrologic forecasting, including recurrent neural networks~\cite{lai2018modeling,siami2018comparison}, hybrid networks~\cite{oreshkin2019n} and graph neural networks~\cite{wu2020connecting,cao2020spectral}. Some work employed Extreme Value Theory (EVT) to enhance the hydrologic time series performance~\cite{li2022extremeadaptive,Yifan2021}. However, these studies primarily concentrate on short-term forecasting and their performance on longer time horizons is doubtful. While there has been a surge in transformer-based forecasting models asserting their high-performance capabilities for long-horizon time series tasks~\cite{li2019enhancing,qin2017dual,zhou2021informer,zhou2022fedformer,kitaev2020reformer}, recent research has raised questions about their efficacy, indicating that simpler linear models can outperform them~\cite{zeng2022transformers,das2023long}. Moreover, imbalanced data or severe events might hurt all these state-of-art deep learning approaches when it comes to long-term predictions. 

We focus on these challenges and innovatively address them through representation learning~\cite{fortuin2018som,lei2017similarity,tonekaboni2022decoupling}, a burgeoning field in unsupervised learning. Our aim is to extract latent states containing the extreme features in data for downstream tasks. To achieve this, we explore the potential of multi-loss functions~\cite{ma2021unsupervised,ma2018modeling} in shaping our training objective. The main contributions of this work are as follows: 
\begin{itemize}
\item We propose a \textbf{D}istance-weighted \textbf{A}uto-regularized \textbf{N}eural network (DAN), which uses expandable blocks to dynamically facilitate long-term prediction.
\item To improve the model's robustness to severe events, DAN innovatively uses a distance-weighted multi-loss method to extract the polar representations from time series simultaneously.
\item We introduce a Kruskal-Wallis sampling policy to handle imbalanced extreme data and gate control vectors to boost the discriminatory capacity of indicators to accommodate imbalanced data.  
\item We evaluate DAN and competing methods on four separate datasets and find that DAN significantly outperforms state-of-the-art baselines. Additionally, we carry out several ablation studies to comprehend the effects of specific design decisions.
\end{itemize}

\section*{Related work}
Streamflow forecasting holds a pivotal role in enhancing water resource allocation, management, flood warning, and mitigation of flood-related damages. Traditional methods for streamflow forecasting included the univariate Autoregressive (AR), Moving Average (MA), Simple Exponential Smoothing (SES), and Extreme Learning Machine (ELM) algorithms, and most famously the Autoregressive Integrated Moving Average (ARIMA)~\cite{boxjen76} method and its several variants. Wang et al.~\cite{w10070853} developed a hybrid model combining Empirical Mode Decomposition (EMD), Ensemble Empirical Mode Decomposition (EEMD) and ARIMA for long-term streamflow forecasting, but they did not examine the effectiveness of their models on datasets with extreme values.

Recently, deep learning models have emerged as the preferred approach for forecasting rich time series data~\cite{sen2019think}, outperforming classical statistical approaches such as ARIMA or GARCH~\cite{box2015time}. Boris et al.~\cite{oreshkin2019n} proposed NBeats, which shows good performance on general time series prediction~\cite{salinas2020deepar}. DeepAR~\cite{salinas2020deepar} learns a conditional distribution over the future values and uses the shared RNN to predict future values and their confidence. To tackle the long-term forecasting challenge, some recent transformer-based methods like Autoformer~\cite{wu2021autoformer} and Reformer~\cite{kitaev2020reformer} have been proposed to empower the transformer with more sophisticated dependency discovery and representation ability. Informer~\cite{zhou2021informer} proposed a ProbSparse self-attention mechanism and a generative style decoder which drastically improves the inference speed of long-sequence predictions. FEDFormer~\cite{zhou2022fedformer} represents time series by randomly selecting a constant number of Fourier components to maintain the global property and statistics of time series as a whole. 

On the other hand, many generic time series prediction models can perform poorly on data with high skewness and kurtosis scores. Conventional methods often falter when confronted with extreme events, which, although infrequent, hold considerable real-world implications—such as in specific instances of streamflow forecasting. Singh et al.~\cite{singh2022credit} proved machine learning approaches suffer from the problem of imbalanced data distribution and noted that balancing the dataset is an imperative sub-task. An and Cho~\cite{an2015variational} proposed an anomaly detection method using the reconstruction probability, which is a probabilistic measure that takes into account the variability of the data distribution. Ding et al.~\cite{ding2019modeling} explored the central theme of improving the ability of deep learning time series models to capture extreme events. Zhang et al.~\cite{Yifan2021} proposed a framework to integrate machine learning models with anomaly detection algorithms. In an earlier work, we proposed NEC+~\cite{li2022extremeadaptive}, a model specifically designed to provide good prediction performance on hydrological time series with extreme events. Additionally, concurrently with this work, we recently developed SEED~\cite{li2023BigData}, a Segment-Expandable Encoder-Decoder model for univariate streamflow prediction.

Few of these prior works have concentrated on addressing both prolonged sequences and extreme occurrences. To bridge this gap, the proposed method, DAN, employs joint learning of two polar hidden spaces within a single series, guided by an associated distance-regularized loss function. This combined approach facilitates accurate end-to-end structure prediction. Remarkably, DAN's performance surpasses that of state-of-the-art methods across four real-life streamflow datasets.
\begin{figure*}[tbh]
\centering
\includegraphics[width=\linewidth]{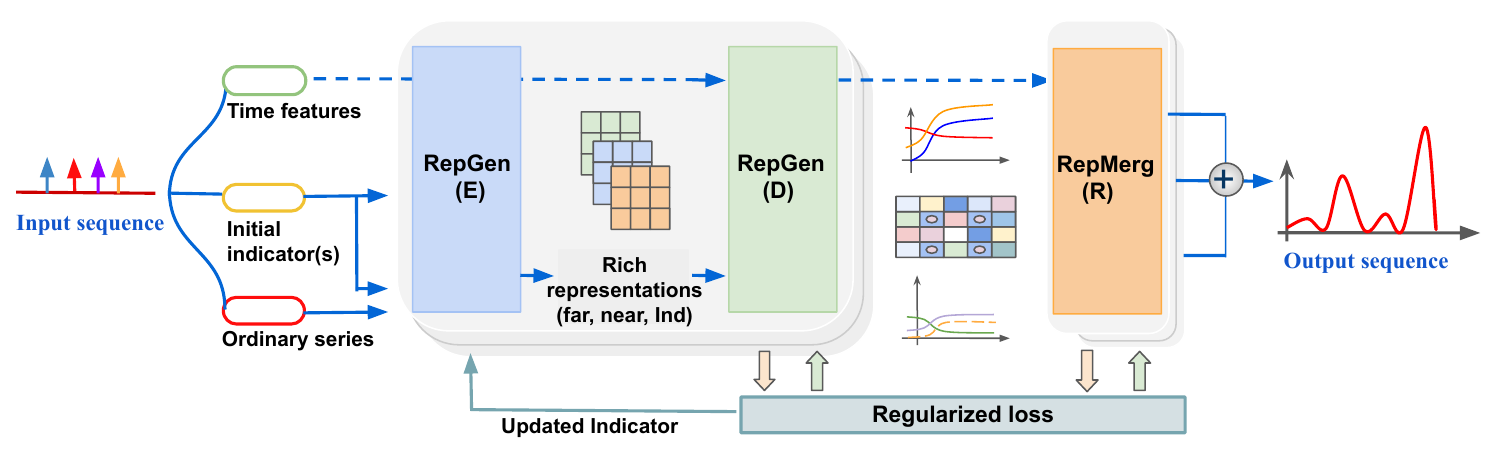}
\caption{DAN's end-to-end extendable framework consists of two stages, named RepGen and RepMerg, respectively. RepGen contains three parallel encoder-decoder blocks, resulting in polar representations of ordinary series inputs and refined indicators. These elements are further merged in the RepMerg stack.}\label{fig:arch}
\end{figure*}

\section*{Preliminaries}\label{sec:prelim}
\subsection{Problem Statement}\label{sec:prelim:problem}
Suppose we have a collection of $m$ ($m>=1$) related univariate time series, with each row corresponding to a different time series. We are going to predict the next $h$ time steps for the first time series $x_1$, given historical data from multiple length-$t$ observed series.  The problem can be described as,
\[
 \begin{bmatrix}
x_{\mathbf{1},1}  & \cdots & x_{\mathbf{1},t}\\
x_{2,1} & \cdots & x_{2,t}\\
\vdots  & \ddots & \vdots\\
x_{m,1}& \cdots & x_{m,t}
\end{bmatrix} \in \mathbb{R}^{m\times t} \rightarrow [x_{\mathbf{1},t+1},  ..., x_{\mathbf{1},t+h}] \in \mathbb{R}^h
\]
where $x_{i,j}$ denotes the value of time series $i$ at time $j$. The matrix on the left are the inputs, and $x_{1,t+1}$ to $x_{1,t+h}$ are the outputs in our method.

We first define this task by modeling the objective time series $x_1$ as the \textit{ordinary series} and the group of related time series $x_2$ to $x_m$ as \textit{extraordinary indicators}. When an extraordinary indicator series is not available, our proposed model can generate a Gaussian Mixture Model (GMM) indicator based on $x_1$, which becomes $x_2$. In such cases, the problem can be reduced to that of univariate time series forecasting.

\subsection{GMM Indicator} \label{sec:prelim:gmm}

In our work, when there is no extraordinary indicator series provided, we use a Gaussian mixture model (GMM)~\cite{day1969} to learn a group of distributions from the input univariate time series. Then, we compute an indicator feature for each value in the time series as the weighted sum of all component probabilities, based on the weights learned during GMM model fitting. Due to lack of space, we detail this step in Section~\ref{sec:appendix:gmm} of our technical appendix.

\subsection{Kruskal-Wallis Test in Time Series}\label{sec:prelim:kw_test}

To balance the sparse distribution of extreme events, we employ the Kruskal-Wallis test~\cite{mckight2010kruskal} as a non-parametric method to evaluate the normality of a training sample and guide our oversampling policy. The Kruskal-Wallis test examines two or more groups of time series based on their medians, in which the data are first ranked, and the sum of ranks is calculated for each group. The $H$ value is then calculated from these rank sums, and compared to a critical value to determine if there are significant differences between the groups. Because the Kruskal-Wallis test does not assume a particular distribution, it is sometimes referred to as a distribution-free test~\cite{ostertagova2014methodology,breslow1970generalized}. The $H$ value is computed as
\begin{equation}\label{eq:kw}
H = \frac{12}{n(n+1)}\sum_{j=1}^k\frac{R_j^2}{n_j}-3(n+1).    
\end{equation}
In our work, we separate the sampled training sequence into $k$ sub sequence groups of equal length. $H$ is the Kruskal-Wallis test statistic, $n$ is the total number of samples across all groups, $R_j$ is the sum of ranks for the $j$th group, and $n_j$ is the number of samples in the $j$th group. The oversampling policy will be described in Section~\ref{sec:methods:sampling}.

\section*{Methods}\label{sec:methods}
An overview of our architecture has been presented in \figurename~\ref{fig:arch}. 

\subsection{Polar Representation Learning}
The key innovations in DAN are new mechanisms to generate and exchange information among polar (\textit{far} and \textit{near}) representations and the indicators (\textit{ind}) for direct improvement of the predictions. Polar representation learning in RepGen allows for the separate encoding of extreme points, which are then preserved in RepMerg. This ensures that, during training, these representations are not compromised by the predominance of normal values, as they often adhere to different distributions. As described in \figurename~\ref{fig:RepMerg}, RepGen consists of three encoder-decoder sub-networks. Please note that the CONV-LSTM layers are specifically designed for learning to predict far points. Because the hidden state for extreme events may be updated multiple times in repeated blocks, we use convolution operations to shorten the input sequence, which helps alleviate any potential exploding or vanishing gradient problem. The \textit{far} and \textit{near} point predictions are extended into the RepMerg stack and further refined as $\hat{y}_{f}$ and $\hat{y}_{n}$, respectively, after being added to the prior output of RepGen. To reflect the change of predicted values, $\hat{y}_{i}$ is expected to converge to the first order of y.

\subsection{Architecture Items}
\begin{figure}[tbh]
\centering
\includegraphics[width=\linewidth]{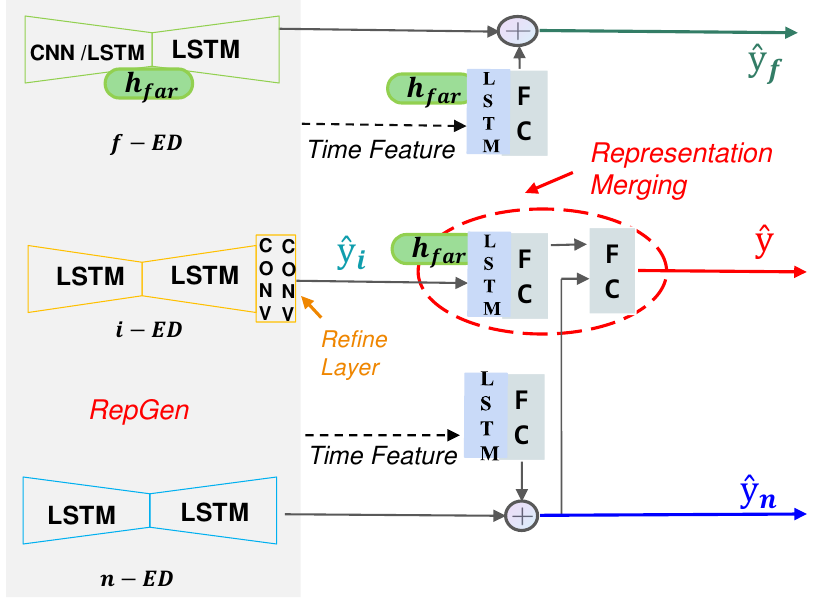}
\caption{ In the RepGen stack, ``\textit{f}-ED'' is responsible for the representation learning of those points that are \textit{far} away from the mean of the series ($\hat{y}_{f}$), including the extreme values sparsely distributed in the predicted zone. ``\textit{n}-ED'' mainly focuses on learning hidden features of \textit{near} points ($\hat{y}_{n}$), which include most of the normal values. ``\textit{i}-ED'' is designed to learn the indicator representation ($\hat{y}_{i}$).}\label{fig:RepMerg}
\end{figure} 

\paragraph{Representation Merge} 
In the RepMerg stack, to finish representation merging, the middle LSTM-FC block in the red circle takes the far point representation $h_{far}$ as the initial hidden state and $\hat{y}_{i}$ as input, which is then combined with $\hat{y}_{n}$ to form a two-dimensional vector as input for another FC layer that outputs $\hat{y}$.

\paragraph{Indicator Refine Layer} 
It should be highlighted that precise $\hat{y}_{i}$  prediction is important for performance improvement as it can help predict extreme events. Therefore, an additional refine layer made of double CNN components is intended to assist in producing the expected indicator by first refining the value within a local horizon. The indicator input can then be cyclically updated from the output of the refine layer of the ``i-ED'' as the RepGen is designed to be an expandable stack that can be repeated multiple times.

\paragraph{Gate Control Vector}
Given the significance of $\hat{y}_{i}$, we provide another way to hone the predicted indicator. As shown in \figurename~\ref{fig:RepMerg}, we designed a gate control vector $m_{far}$, whose values reflect precisely in which places $\hat{y}$ is closer to $\hat{y}_{f}$. Similarly, $m_{near}$ is the complement of $m_{far}$. We then compute $\hat{y}_{w}$ as $m_{far} \odot \hat{y}_{f} + m_{near} \odot \hat{y}_{n}$, where $\odot$ is the component-wise multiplication. We therefore use $\hat{y}_{w}$ to increase the discriminatory ability of $\hat{y}_{i}$ when it is expected to reflect better extreme values without sacrificing the overall normal values.
\begin{figure}[htb]
\centering
\includegraphics[width=\linewidth]{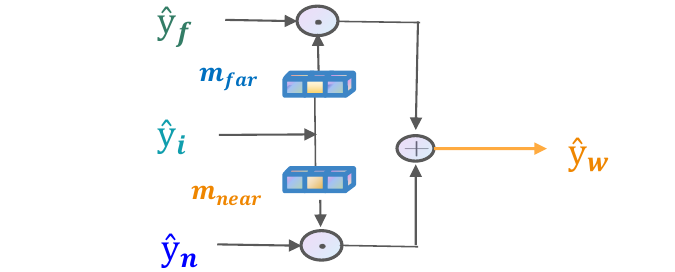}
\caption{Gate control vector. $m_{far}$ is equal to $sigmoid(\alpha \hat{y}_{i})$, where $\alpha$ is an amplifier and equals $4$ in our experiments; $m_{near}$ is computed as $1-m_{far}$.}.\label{fig:gate}
\end{figure}
\subsection{Auto-Regularized Loss Function}\label{sec:methods:loss}
Different from the conventional usage of regularization loss as a penalty term for preventing overfitting of the model to the training data, our approach employs multiple distance-weighted loss functions when training the DAN model, with the objective of compelling the model to learn more informative representations. Moreover, it should be noted that our method can also serve as an effective regularizer for preventing overfitting of the model to the base normal values in the long-term time series prediction task.

We define $w_{f} = (\tanh(y))^2 $ to be a weight that emphasizes the accuracy of points further away from the mean value of 0 (since the series were standardized to have 0 mean). In contrast, $w_{n} = (1 - |\tanh(y)|)^2$ focuses more on the points closer to zero. The square root of $w_{f}$, denoted by $w_{p}$, is a more moderate way to maintain discriminatory output for indication-related tasks. These weight definitions contribute to the accuracy of the model in predicting extreme events in a long-term time series. Based on these weights, we build our multi regulation loss as follows,
\begin{align*}
\mathcal{L}_1 &= RMSE(( \hat{y}_{f} \odot w_{f}), (y \odot w_{f})),\\
\mathcal{L}_2 &= RMSE(( \hat{y}_{n} \odot w_{n}), (y \odot w_{n})),\\
\mathcal{L}_3 &= RMSE( \hat{y}_{w}  \odot w_{p}, y \odot w_{p}),\\
\mathcal{L}_4 &= RMSE( \hat{y}_{i}  \odot w_{p}, y_{i}\odot w_{p}),
\end{align*}
where $\mathcal{L}_1$ and $\mathcal{L}_2$ are used to regulate the bipolar representation learning and  $\mathcal{L}_3$ and $\mathcal{L}_4$ force the predicted indicator to reflect the change of predicted values by setting $y_{i}$ equal to the first order of $y$. Then, the overall loss is composed as,
\[
\mathcal{L}  = RMSE(\hat{y},  y) + \lambda \times (\mathcal{L}_1 + \mathcal{L}_2 +\mathcal{L}_3 + \mathcal{L}_4),
\]
where $\lambda$ is a multiplier ($\lambda = max( -1 \cdot e^{\frac{epoch}{45}} + 2, 0.2)$ in our experiments) applied on those regulation items, decreasing with each epoch.  

\subsection{Kruskal-Wallis Sampling}
\label{sec:methods:sampling} 
Given that extreme events are rare within our data compared with normal ones, we utilize Kruskal-Wallis sampling to over-sample regions with extreme events in our training set that our model can learn appropriate patterns from. Namely, for each random sample $x$ of size $t+h$ we draw from the input sequence, we first split the sequence into $k$ consecutive sub-sequences of equal size and compute the Kruskal-Wallis test statistic $H$ between the $k$ sub-sequences, using Equation~\ref{eq:kw}. To avoid the $H$ statistic being affected by minor differences in the sub-sequences, we round values in $x$ to the nearest integer before computing $H$. We then include the sample in the training set if $H > \epsilon$, where $\epsilon$ is a threshold, or otherwise include the sample with probability $p < 1$. The threshold $\epsilon$ allows us to set the relative change in the sample that makes it more likely to contain an extreme event, while the probability $p$ allows us to choose how many normal samples should be included in the training set.

\section*{Evaluation}\label{sec:experiments}
Code and data for DAN can be found at~\url{https://github.com/davidanastasiu/dan}. In this section, we present empirical results from evaluating our proposed framework. We are interested in answering the following research questions with regards to prediction effectiveness: (1) How does DAN compare against state-of-the-art baselines? (2) What is the effect of DAN’s extensible framework? (3) What is the effect of the Kruskal-Wallis oversampling policy? (4) How do the critical design elements of the framework affect performance?

\subsection{Experimental Settings}\label{sec:experiments:design}

\paragraph{Dataset} We used four groups of hydrologic datasets from Santa Clara County, CA, namely Ross, Saratoga, UpperPen, and SFC, named after their respective locations. Each group included a streamflow dataset and an associated rainfall dataset. Statistics of our primary series are shown in Table~\ref{tab:dataset_stats}. Our task was to forecast the streamflow for wet seasons in a hydrologic year, excluding the summer months, namely September 2021 to May 2022, in a rolling manner. The training and validation sets were randomly sampled from series spanning from January 1988 to August 2021. Inference involved predicting the streamflow every 4 hours for the next 3 days. Since the sensors measure the streamflow and precipitation every 15 minutes, we are attempting a lengthy forecasting horizon ($h=288$), which is unquestionably an LSTF task based on the most recent research~\cite{zhou2021informer, zhou2022fedformer, zeng2022transformers,das2023long}. Before training a model, all time series were pre-processed by log transform ($x_i = \log(1+x_i) \forall i$) and standardization (subtract mean and divide by standard deviation). Inference predictions were post-processed by inverting the standardization and log transform operations.

\begin{table}[tbh]
\center{
\caption{Streamflow Datasets Statistics}\label{tab:dataset_stats}
    \begin{tabular}{crrrr}
    \cmidrule(r){1-5}
         & Ross & Saratoga & UpperPen & SFC  \\ 
         \cmidrule(r){1-5}
        min & 0.00 & 0.00 & 0.00 & 0.00  \\ 
        max & 1440.00 & 2210.00 & 830.00 & 7200.00  \\ 
        mean & 2.91 & 5.77 & 6.66 & 20.25  \\ 
        skewness & 19.84 & 19.50 & 13.42 & 18.05  \\ 
        kurtosis & 523.16 & 697.78 & 262.18 & 555.18  \\ \cmidrule(r){1-5}
    \end{tabular}
}    
\raggedright 
{\footnotesize High skewness and kurtosis scores indicate that there is a significant deviation from a normal distribution in our data.}
\end{table}

\paragraph{Model Parameters} For all models, after testing different LSTM layer widths, we found that 512 node layers for ROSS and 384 layers for the other three sensors were the most effective. We set $h = 288$ (3 days) based on the problem definition and tested different values of $t \in \{288, 672, 1440\}$ (3, 7, 15 days), with $t = 1440$ producing the best results for all data streams. In the RepGen stage, the three CNN layers produce 256 channels each. The kernel sizes used in these layers are 11, 7, and 3, respectively. The stride, padding, and activation function remain the same across all three layers, with a stride equal to the kernel size, no padding, and a subsequent $tanh$ activation function. We use two stacked CNN1d layers for indicator refinement, with the kernel size and padding set to 7 for the first layer and 3 for the second. 

\subsection{Experimental Results}\label{sec:experiments:results}
\paragraph{Baselines}
We include seven state-of-the-art models for comparison, of which FEDFormer~\cite{zhou2022fedformer}, Informer~\cite{zhou2021informer}, NLinear~\cite{zeng2022transformers}, and DLinear~\cite{zeng2022transformers} focus on long-term time series forecasting, while NEC+~\cite{li2022extremeadaptive} holds the best performance for hydrologic time series prediction in the presence of extreme events. These five models were used as baselines for both multivariate and univariate prediction. In addition, Attention-LSTM~\cite{le2021attention} was used as a state-of-the-art hydrologic multivariate model used to predict stream-flow using rainfall data. Finally, N-BEATS~\cite{oreshkin2019n}, a state-of-the-art univariate baseline method that outperformed all competitors on the standard M3~\cite{MAKRIDAKIS2000451}, M4~\cite{makridakis2018m4}, and TOURISM~\cite{athanasopoulos2011tourism} time series datasets, was also used in the comparison.

\begin{table*}[ht]
\vspace{0.5em}
\center {
\caption{Multivariate/Univariate Long-Term ($h=288$) Series Forecasting Results on Four Datasets}
    \label{tab:results}

    \resizebox{0.8\textwidth}{!}{
    \begin{tabular}{ccrrrrrrrr}
    \toprule
        \textbf{Methods} & \textbf{Metric} &  \multicolumn{2}{c}{\textbf{Ross}}&  \multicolumn{2}{c}{\textbf{Saratoga}} &  \multicolumn{2}{c}{\textbf{UpperPen}} & \multicolumn{2}{c}{\textbf{SFC}}  \\ 
        \cmidrule(r){3-10}
        \textbf{} & ~ & \multicolumn{1}{c}{Multi} & \multicolumn{1}{c}{Single} & \multicolumn{1}{c}{Multi} & \multicolumn{1}{c}{Single} & \multicolumn{1}{c}{Multi} & \multicolumn{1}{c}{Single} & \multicolumn{1}{c}{Multi} & \multicolumn{1}{c}{Single}  \\ 
        \cmidrule(r){1-10}

        \textbf{FEDformer} & RMSE & \underline{6.01}  & 6.49  & 6.01  & 6.85  & 3.05  & 2.38   &  23.54  & 24.10 \\
        \textbf{} & MAPE & 2.10 &  2.49 & 1.55 &  2.26 & 1.87 & 1.02 &      2.35 &       2.817 \\
        \cmidrule(r){1-10}
        \textbf{Informer} & RMSE & 7.84  & 9.14   & 5.04  & 4.89  & 5.88  &5.33  &     39.89  &   19.00  \\
        \textbf{} & MAPE & 4.05 &  5.45 & 1.43 &  0.73 & 4.10 & 4.21 &      8.64 &      \underline{0.54} \\
        \cmidrule(r){1-10}
        \textbf{Nlinear} & RMSE & 6.10   & 5.84   & 5.23  & 4.98 & \underline{1.57}  & 1.74  &     18.47  &    18.43   \\
        \textbf{} & MAPE & \underline{1.99} &  1.62 & 0.83 &  0.75 & \underline{0.45} & 0.57 & \underline{0.92} &  0.87 \\
        \cmidrule(r){1-10}
        \textbf{Dlinear} & RMSE & 7.16  & 6.90  & 4.33  & 4.06  & 3.53  & 3.25  &     21.62  &     23.64 \\
        \textbf{} & MAPE & 3.10   & 2.79 & 1.40 &  1.31 & 2.35 &  2.04 &      2.74 &   4.02 \\
        \cmidrule(r){1-10}
        \textbf{NEC+} & RMSE & 9.44  & 9.33  & \underline{1.88}  &  \underline{1.95}   & 2.22  & 1.94   &     \underline{17.00}     &    \underline{16.39}    \\
        \textbf{} & MAPE & 4.80 &  4.53  & \underline{0.17} &  \underline{0.21} & 0.95 & 0.80 &      1.07 &  0.55  \\
        \cmidrule(r){1-10}
        \textbf{LSTM-Atten /} & RMSE & 7.35  & \underline{5.16} & 6.49  & 3.60  & 6.35  & \textbf{1.23}   &  34.17  & 31.47   \\
        \textbf{NBeats} & MAPE & 3.74 &  \underline{1.25} & 1.80 & 0.70& 4.76 & \textbf{0.25} &      9.90 &     3.24\\
        \cmidrule(r){1-10}
        \textbf{DAN} & RMSE &\textbf{4.25}  & \textbf{4.24}  & \textbf{1.80}  &   \textbf{1.84} & \textbf{1.10}  &  \underline{1.31} &   \textbf{15.23}   &   \textbf{15.20} \\
        \textbf{} & MAPE & \textbf{0.07} & \textbf{0.09} & \textbf{0.14} &  \textbf{0.16}  & \textbf{0.15} &  \underline{0.32}  &  \textbf{0.26} & \textbf{0.21}\\
        \bottomrule            \end{tabular}} 
}

\raggedright
\noindent
{\footnotesize Over 1600 test points in the test set were inferenced on all datasets. The best results are in bold and the second best results are underlined.}
\end{table*}

\paragraph{Multivariate and Univariate Results}

\begin{figure*}[tbh]
  \centering {\includegraphics[width=\linewidth]{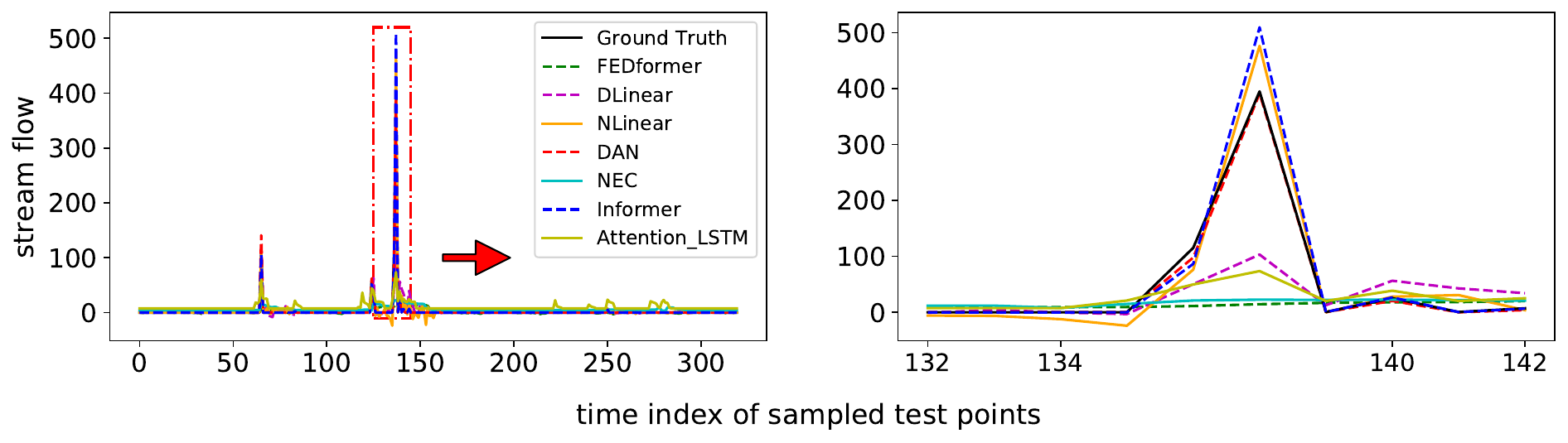}} 
\caption{Sampled multivariate inference for the Ross sensor. The right figure emphasizes extreme events occurring during the example time period in the red box of the left figure.}\label{fig:YearFLow}
\end{figure*}

Table~\ref{tab:results} shows the test root mean squared error (RMSE) and mean absolute percentage error (MAPE) performance for the models that achieved the best performance on our validation dataset. For these experiments, all DAN results were achieved using the same random seed. Due to lack of space, we include the definition of our performance metrics and multiple seed run performance statistics in Section~\ref{sec:appendix:experiments} of our technical appendix. In the multivariate forecasting task, our proposed model DAN outperformed all baselines on all four benchmark datasets. Compared to the second-best results, DAN achieved an overall 19\% relative RMSE reduction. Notably, the improvement was most significant for the UpperPen dataset, where DAN achieved a 30\% improvement. For the univariate forecasting task, DAN outperformed other methods on three out of four benchmark datasets. Although NBeats achieved a 6\% relative RMSE reduction compared to DAN for the UpperPen sensor, DAN surpassed NBeats with a relative RMSE reduction of 18\%, 49\%, and 52\% on the Ross, Saratoga, and SFC datasets, respectively. 

\paragraph{Inference Overall Analysis}
\figurename~\ref{fig:YearFLow}, in which we present rolling prediction results for the whole test set (1600 time points) for the Ross sensor, helps explain DAN's good performance. To make it easier to visualize, the test set was sampled every 5 points (320 time points sampled) and a specific period including extreme events is denoted by the red box and focused on in the right figure. We observed that DAN performed better than other models on areas with extreme events. While Informer and NLinear could follow the extreme events better than other baselines, they predicted values much higher than the actual peaks. On the other hand, DLinear and Attention-LSTM performed better than NEC+ and FEDFormer, but they predicted values much lower than the ground truth. 

\subsection{Ablation Studies}\label{sec:experiments:ablation}
The superior performance of our method comes from its extendable framework, creative Kruskal-Wallis sampling policy, and three key architecture designs: 1) the use of a refine layer in RepGen that refines the indicator information using double CNNs with moving kernel convolution operations, 2) the RepMerg layer which combines the rich polar representations with the indicator learned from RepGen, and 3) regularization loss components. We will examine their effects respectively in this section.
\begin{figure*}[tbh]
\centering
\includegraphics[width=\linewidth]{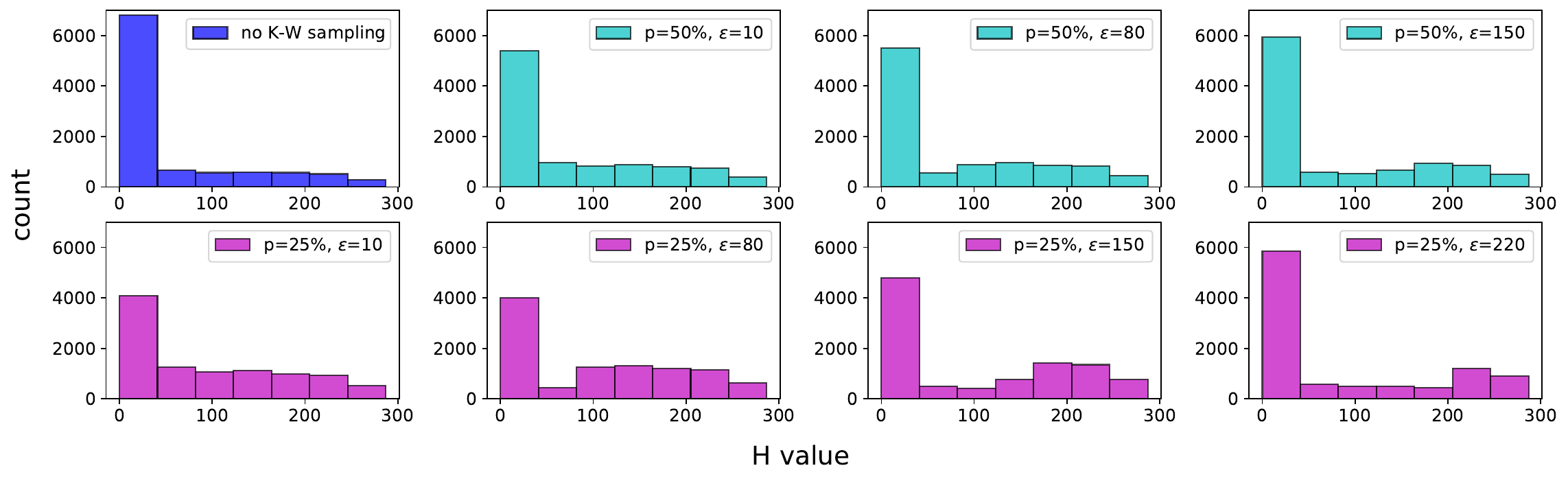}
\caption{Oversampling policy. Initially, a majority of the samples have $H$ values below 10. However, upon setting $p=50\%$ and $\epsilon=10$, the number of samples with $H$ values below 10 decreases, while the count for higher $H$ values increases. The second row showcases the scenario for $p=25\%$. }\label{fig:oversampling}
\end{figure*}
\subsubsection{Effects of DAN's Extensible Framework}

Our network comprises two main stages. The repeatable stacks are named ``E'', ``D'' and ``R'' in \figurename~\ref{fig:arch}, endowing DAN with the capability of updating the indicator in an evolutional way by repeating ``E''+``D'' and including multiple refinements by repeating ``R''. The best stack configuration may vary for different datasets, depending on the intrinsic relationships in the multivariate series. We experimented with various combinations and identified the best results as ``EDEDRR'', ``EDR'', ``EDEDRR'', and ``EDEDR'' for Ross, Saratoga, UpperPen, and SFC, respectively. 

\subsubsection{Effects of the Oversampling Policy}

\figurename~\ref{fig:oversampling} displays the distribution of the $H$ values before and after applying the Kruskal-Wallis sampling algorithm.  These observations highlight the impact of adjusting $p$ and $\epsilon$ on the distribution of $H$ values in the training set. If we maintain the $p$ value and increase the $\epsilon$ value, the training set will contain more samples with $H$ values exceeding $\epsilon$, as illustrated in the rightmost three figures in the first row of \figurename~\ref{fig:oversampling}. 

\begin{table}[htb]
\center{
  \caption{$RMSE_{far}$ when Oversampling}
  \label{tab:oversampling_policy}

  \begin{tabular}{ccc}
    \toprule
    Policy($\epsilon=10$) & Ross & SFC  \\
    \cmidrule(r){1-3}
    $20\%<=p<=40\%$  & \textbf{23.5} & \textbf{118.1}\\
    \cmidrule(r){1-3}
    $60\%<=p<=80\%$  & 25.0 & 122.0 \\
    \cmidrule(r){1-3}
    $p=100\%$  & 28.7  & 129.0\\
    \bottomrule
  \end{tabular}
}

\raggedright
\noindent
{\footnotesize We used $\epsilon$ = 10 and the $p$ values used were grouped into 3 cases. $p=100\%$ means no Kruskal-Wallis oversampling was applied. }
\end{table}

Our oversampling policy can help shift the focus of our model towards improving ``far'' point prediction performance. To test this, we conducted multiple runs of each model, averaging the RMSE values for points with values greater than 1.5 standard deviations above the mean of the series, which we call $RMSE_{far}$ in Table~\ref{tab:oversampling_policy}. The results show that the $RMSE_{far}$ can be steadily decreased by decreasing $p$. 

\subsubsection{Effects of DAN Architecture Elements}

\begin{figure}[bht]
\centering
\includegraphics[width=\linewidth]{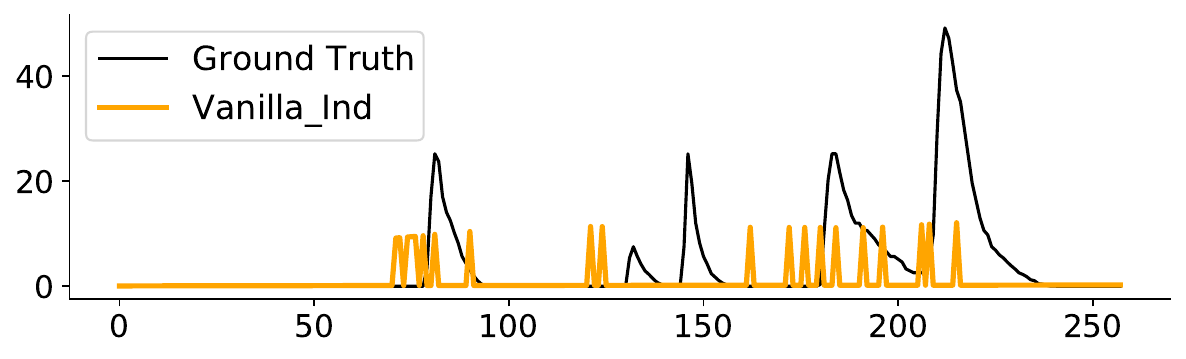}
\includegraphics[width=\linewidth]{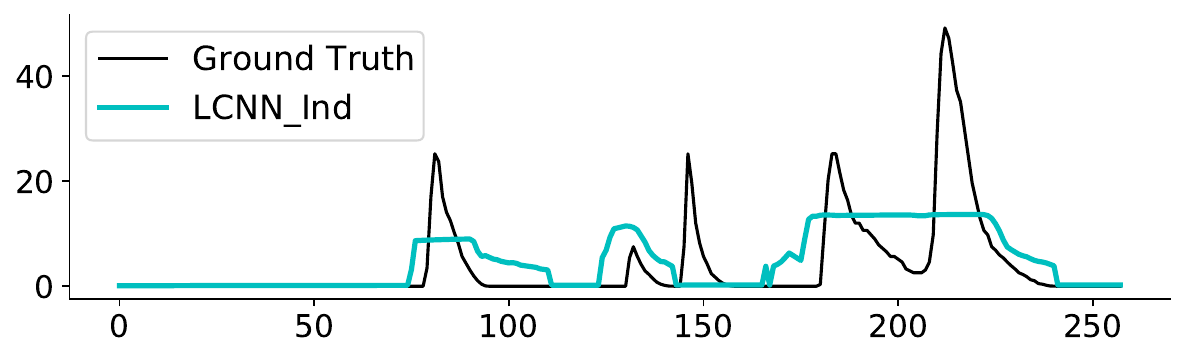}
\includegraphics[width=\linewidth]{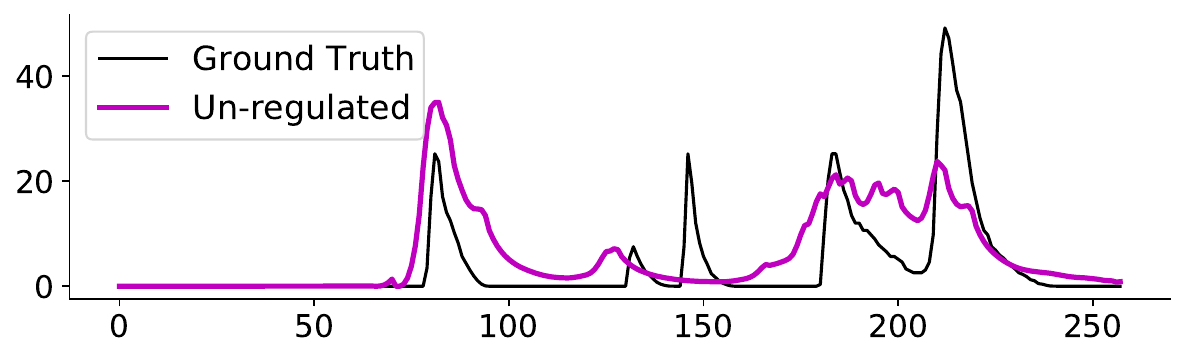}
\includegraphics[width=\linewidth]{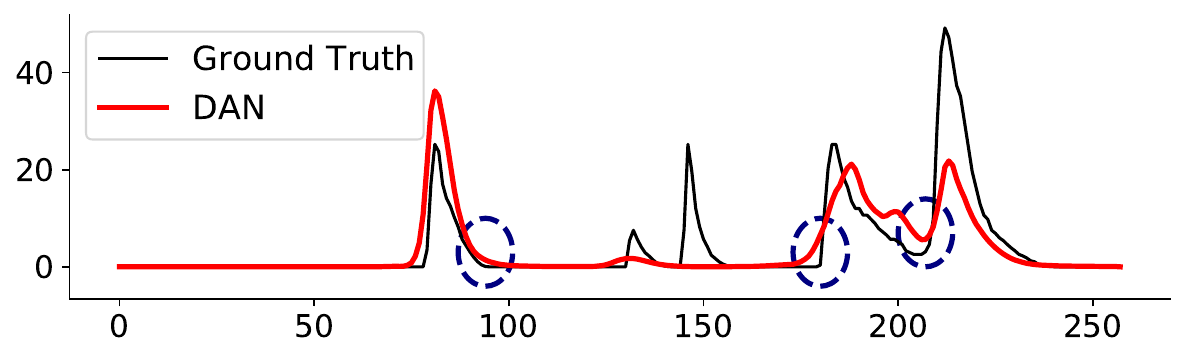}
\caption{Inference examples to show the effects of different architecture elements on the Ross dataset.}\label{fig:comp}
\end{figure}

In this section, we investigate the impact of different design elements on the performance of our network. To isolate the effects of these design elements and obtain a more comprehensive understanding of each one, we replace the input indicator in our network with ground truth rain data and only use a simple EDR architecture stack, which will remove the effect of varying indicator performance from the comparison. By doing so, we can observe how the network performs when certain layers are removed or added back.

To create a baseline structure, we first removed the refine and the RepMerge layers, using only the encoder-decoder block to generate the indicator, and set $\hat{y}$ equals to $\hat{y}_{w}$ to bypass the RepMerge structure. This produced the first predicted sequence (named \textbf{Vanilla\_Ind} in \figurename~\ref{fig:comp}), which shows an example of inference on Ross sensor data. We then added the refine layer back, which improved the results as shown in the second sub-figure (named \textbf{LCNN\_Ind}). Next, we added back the RepMerg layer.
Concurrently, we also removed the regularization loss items except $\mathcal{L}_3$, obtaining the third figure (named \textbf{Un-regulated}). Finally, we added all regularization loss items back, which gave the best result, as shown in the fourth figure. 

These experiments demonstrate that the double CNNs with moving kernel convolutional operations refines the indicator information, and RepMerg produces better results as the Gate control vector mechanism increases the discrimination of predicted values. Adding polar representation of the basic series assists in identifying data distributions beyond the indicator information and enhances the accuracy of data at corners of each fluctuation, as denoted in the blue circles in \figurename~\ref{fig:comp}. Therefore, including the refine layer, representation learning, and gate control vector resulted in the best performance.

\section*{Conclusion}
In this work, we presented a novel end-to-end framework, DAN, designed to better account for rare yet important extreme events in long single- and multi-variate streamflow time series. Our framework learns polar representations for predicting extreme and normal values, along with a representation merging model that makes prediction in an expandable way. In addition, to improve training performance, our framework uses Kruskal-Wallis sampling policies to accommodate imbalanced extreme data and a distance-weighted multi-loss regularization penalty. Extensive experiments using more than 33 years of streamflow data from Santa Clara County, CA, showed that DAN provides significantly better predictions than state-of-the-art baselines.


\section*{Appendix}

\section{Network description}\label{sec:appendix:network}
\begin{figure*}[tbh]
  \centering {\includegraphics[width=0.8\linewidth]{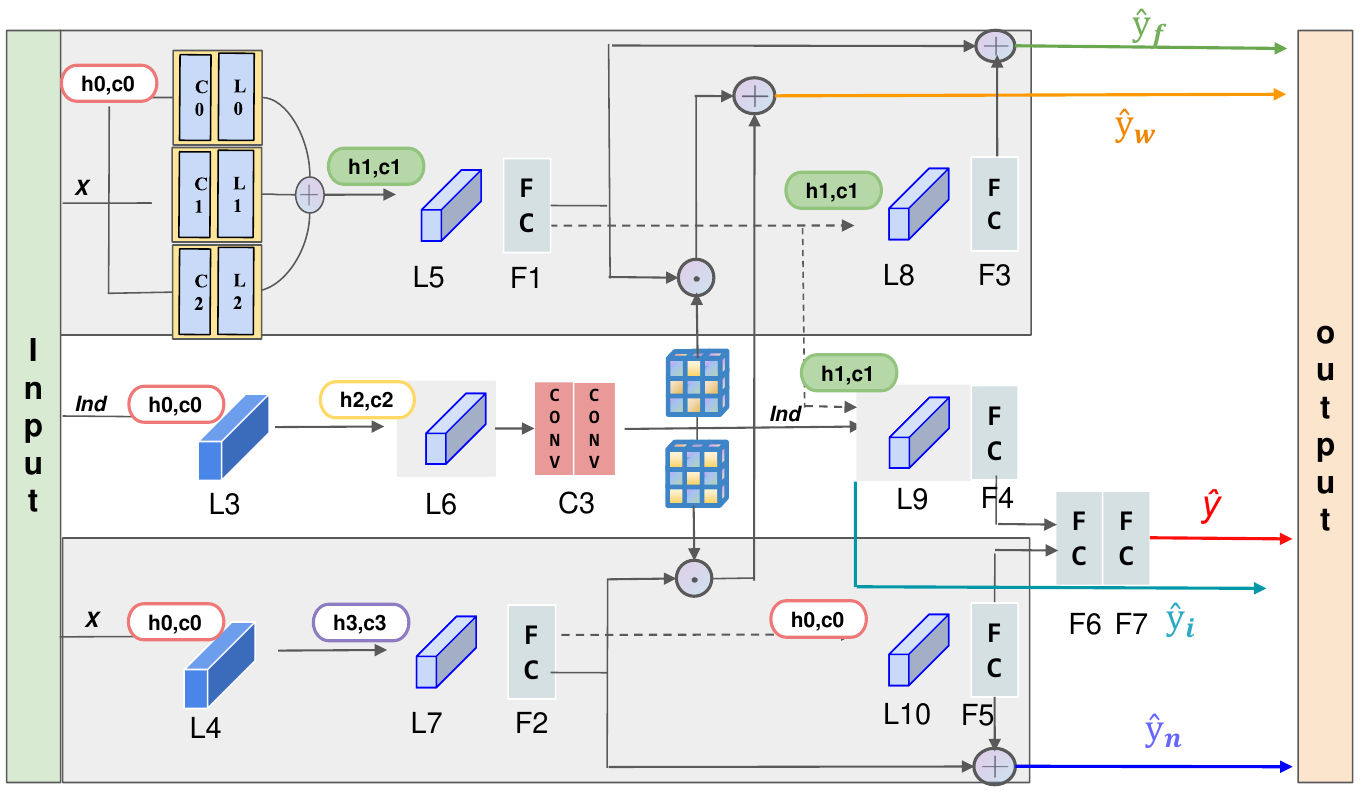}} 
\caption{Data flow and architecture layers.}\label{fig:dataflow}
\end{figure*}

For all models, we used 10 bidirectional LSTMs with 512 hidden features for ROSS and 384 for the others. Figure~\ref{fig:dataflow} shows a detailed representation of our framework.

In the RepGen stage, the three CNN layers are labelled as C0, C1 and C2. The stride, padding, and activation functions are listed in Table~\ref{tbl:conv_layers}. Two stacked CNN1d layers are used to finish the indicator refinement, which we noted as C3. Each layer has 256 channels with a stride of 1. The kernel size and padding for the first two layers are set to 7 and 3, respectively. All the fully connected layers have $384$ (512 for ROSS)$\times 2$ input features and 1 output feature.
 
\begin{table*}[!ht]
\caption{Convolution Layers in Our Framework}
\label{tbl:conv_layers}
    \centering
    \begin{tabular}{ccccccc}
    \hline
        layer & in\_dim & out\_dim & kernel size & stride & padding & acti func  \\ \hline
        C0 & 1 & 256 & 11 & 11 & 0 & tanh  \\ 
        C1 & 1 & 256 & 7 & 7 & 0 & tanh  \\ 
        C2 & 1 & 256 & 3 & 3 & 0 & tanh  \\ 
        \multirow{2}{*}{C3} & $512/384\times 2$ & 512/384 & 7 & 1 & 3 & \multirow{2}{*}{relu}  \\ 
        ~ & 512/384 & 1 & 3 & 1 & 1 &   \\ \hline
    \end{tabular}
\end{table*}

\begin{table}[!ht]
\caption{LSTM Layers in Our Framework}
    \centering
    \begin{tabular}{cccccc}
    \hline
        layer & in\_dim & out\_dim & layer &  bidirectional  \\ \hline
        L1 & 256 & 512/384 & 1  & yes  \\ 
        L2 & 256 & 512/384 & 1 & yes  \\ 
        L3 & 1 & 512/384 & 1 &  yes  \\ 
        L4 & 1 & 512/384 & 1 &  yes  \\ 
        L5 & 2 & 512/384 & 1 & yes  \\ 
        L6 & 2 & 512/384 & 1 & yes  \\ 
        L7 & 2 & 512/384 & 1 &  yes  \\ 
        L8 & 2 & 512/384 & 1 & yes  \\ 
        L9 & 1 & 512/384 & 1 & yes  \\ 
        L10 & 2 & 512/384 & 1  & yes  \\ \hline
    \end{tabular}
\end{table}

\section{Dataset description}\label{sec:appendix:stats}
To analyze our data set, we computed the related statistical values of our input time series. We use the measures of skewness and kurtosis to provide valuable insights into the shape and distribution of the data. The high skewness and kurtosis shows that the datasets deviate significantly from a normal distribution. In our datasets, the positive skewness reflects the asymmetry of the distribution. To be specific, it indicates a longer right tail. On the other hand, the kurtosis values in the hundreds indicate heavier tails and a higher peak. 

\begin{table*}[!ht]
\caption{Statistics of our Stream Data}
\label{tbl:dataset_stats}
    \centering
    \begin{tabular}{ccccc}
    \hline
        Statistic / Stream & Ross & Saratoga & UpperPen & SFC  \\ \hline
        mean & 2.91 & 5.77 & 6.66 & 20.25  \\ 
        max & 1440.00 & 2210.00 & 830.00 & 7200.00  \\ 
        min & 0.00 & 0.00 & 0.00 & 0.00  \\ 
        median & 0.17 & 1.00 & 3.20 & 1.20  \\ 
        variance & 597.22 & 711.09 & 452.90 & 12108.14  \\ 
        std\_deviation & 24.43 & 26.66 & 21.28 & 110.03  \\ 
        skewness & 19.84 & 19.50 & 13.42 & 18.05  \\ 
        kurtosis & 523.16 & 697.78 & 262.18 & 555.18  \\ \hline
    \end{tabular}
\end{table*}

Table~\ref{tbl:dataset_stats} shows the computed statistics of our input time series. From these data, we can find these data sets have a pronounced departure from normality, with a significant skew and heavy tails. This implies that the distribution of the data is skewed towards one side and contains extreme values that deviate from the average.
\section{Related research}\label{sec:appendix:related}
In this section, we discuss several studies that have employed traditional machine learning methods for stream flow prediction. Nayak et al.~\cite{NAYAK200452} utilized neural network models (NN) and adaptive neuro-fuzzy inference systems (ANFIS) to forecast stream flow. Tabbussum et al~\cite{tabbussum2021comparison} proposed a fuzzy inference system termed as flood model using two fuzzy flood models. Gaussian process regression (GPR) and quantile regression were used in some studies to not only predict but also quantify forecast uncertainty. Tree-based models, such as classification and regression trees (CARTs) and random forest (RF), have been employed due to their computational efficiency and ability to handle predictors without assuming any specific distribution. Nhu et al.~\cite{RFbased} demonstrated the use of CART and RF in solving hydrologic prediction problems.  While neural networks have been utilized in early hydrologic flow prediction studies, they were often shallow networks unable to capture complex patterns in the data, requiring extensive feature engineering and manual tuning based on domain expertise to improve performance.

\section{GMM Indicator Generation}\label{sec:appendix:gmm}

A Gaussian mixture model (GMM)~\cite{day1969-a} can be described by the equation,
\[
p(x|\lambda)=\sum_{i=1}^M w_i~g(x|\mu_i,\mathbf{\Sigma_i}),
\]
where $x$ is a $D$-dimensional continuous-valued vector, $w_i\ \forall i = 1, \ldots, M$ are the mixture weights, and $g(x|\mu_i, \Sigma_i)$, are the component Gaussian densities. Each component density is a $D$-variate Gaussian function, and the overall GMM model is a weighted sum of M learnable Gaussian densities, i.e.,
\begin{footnotesize}
$$
g(x|\mathbf{\mu_i},\mathbf{\Sigma_i})=\frac{1}{2\pi^\frac{D}{2}|\Sigma_i|^\frac{1}{2}}\exp{\left\{-\frac{1}{2}{(x-\mu_i)}^T~ \Sigma_i^{-1} (x-\mu_i)\right\}},
$$
\end{footnotesize}
where $\mu_i$ is the mean vector and $\Sigma_i$ is the covariance matrix of the $i$th component. The mixture weights are constrained such that $\sum_{i=1}^M w_i=1$. 
The GMM's capacity to produce smooth approximations to arbitrarily shaped densities is one of its most impressive features. 

In our work, we utilize the Expectation-Maximization (EM) algorithm to fit a Gaussian Mixture Model (GMM) to the ordinary series training data. The GMM consists of multiple components, and each component represents a Gaussian distribution. The EM algorithm iteratively estimates the parameters of the GMM by maximizing the likelihood of the observed data.

Once we have fitted the GMM, each model component can generate a probability for each point in the time series. These probabilities represent the likelihood that a particular component generated the observed data point. To further analyze the time series, we compute an "indicator" feature for each value in the time series. The indicator feature is computed as the weighted sum of all component probabilities, given the weights learned during the fitting process of the GMM model. This feature provides a measure of the contribution of each component to the overall time series at a specific point.

In our framework, the number of components in the GMM, denoted as $M$, is a hyper-parameter that can be tuned for each time series. We set it as 3 in all our experiments. 

\section{Experiments}\label{sec:appendix:experiments}
\subsection{Evaluation metrics}
We use the popular Root Mean Square Error (RMSE) and Mean Absolute Percentage Error (MAPE) metrics to measure the effectiveness of our prediction methods. RMSE is defined as the square root of the mean squared error between the predicted and actual values, while MAPE is the mean of all absolute percentage errors between the predicted and actual values.
\subsection{Multiple seed performance statistics}
\begin{table*}[bht]
\caption{Multiple seed performance statistics}
    \centering
    \label{tab: multiple_run}
    \begin{tabular}{ccccccccc}
    \hline
        \multirow{2}{*}{\textbf{RMSE}} &\multicolumn{2}{c} {\textbf{Ross}} & \multicolumn{2}{c}{\textbf{Saratoga}} & \multicolumn{2}{c}{\textbf{UpperPen}} & \multicolumn{2}{c}{\textbf{SFC}}  \\ \cmidrule(r){2-9}
         & multi & single & multi & single & multi & single & multi & single  \\ \cmidrule(r){1-9}
        min & 4.24 & 4.23 & 1.71 & 1.82 & 1.03 & 1.12 & 14.99 & 15.20  \\ \cmidrule(r){1-9}
        max & 4.36 & 4.35 & 1.96 & 2.06 & 1.35 & 1.31 & 15.86 & 17.56  \\ \cmidrule(r){1-9}
        mean & \textbf{4.29} & \textbf{4.31} & \textbf{1.84} & \textbf{1.91} & \textbf{1.18} & \textbf{1.19} & \textbf{15.55} & \textbf{16.34}  \\ \cmidrule(r){1-9}
        std\_deviation & 0.04 & 0.02 & 0.08 & 0.09 & 0.12 & 0.06 & 0.26 & 0.55  \\ \hline
    \end{tabular}
\end{table*}

In this part of the study, we aim to mitigate the impact of randomness and obtain a more comprehensive evaluation by training our models multiple times with different random seeds. Specifically, we performed 6 to 8 runs for the multivariate models and the univariate models. The resulting minimum, maximum, mean, and standard deviation values are summarized in the Table~\ref{tab: multiple_run}. The oversampling parameters used in this experiment are $p=25\%, 60\%, 18\%, 18\%$ and $\epsilon=30, 150, 80, 80$ for the Ross, Saratoga, UpperPen, and SFC streams, respectively.

Despite the average RMSE being slightly higher than the minimum values, it is important to note that DAN consistently outperformed all baselines across all four benchmark datasets. 


\begin{acknowledgements}
Funding for this project was made possible by the Santa Clara Valley Water District.
\end{acknowledgements}

\section*{Bibliography}
\bibliography{arxiv.bib}

\begin{thebibliography}{49}
\providecommand{\natexlab}[1]{#1}
\providecommand{\url}[1]{\texttt{#1}}
\expandafter\ifx\csname urlstyle\endcsname\relax
  \providecommand{\doi}[1]{doi: #1}\else
  \providecommand{\doi}{doi: \begingroup \urlstyle{rm}\Url}\fi

\bibitem[Nguyen and Chan(2004)]{nguyen2004multiple}
Hanh~H Nguyen and Christine~W Chan.
\newblock Multiple neural networks for a long term time series forecast.
\newblock \emph{Neural Computing \& Applications}, 13:\penalty0 90--98, 2004.

\bibitem[Ding et~al.(2019)Ding, Zhang, Pan, Yang, and He]{ding2019modeling}
Daizong Ding, Mi~Zhang, Xudong Pan, Min Yang, and Xiangnan He.
\newblock Modeling extreme events in time series prediction.
\newblock In \emph{Proceedings of the 25th ACM SIGKDD International Conference
  on Knowledge Discovery \& Data Mining}, pages 1114--1122, 2019.

\bibitem[Qi and Majda(2020)]{qi2020using}
Di~Qi and Andrew~J Majda.
\newblock Using machine learning to predict extreme events in complex systems.
\newblock \emph{Proceedings of the National Academy of Sciences}, 117\penalty0
  (1):\penalty0 52--59, 2020.

\bibitem[Chen et~al.(2020)Chen, Yu, Geng, Li, and Zhang]{chen2020evanet}
Zechuan Chen, Haomin Yu, Yangli-ao Geng, Qingyong Li, and Yingjun Zhang.
\newblock Evanet: An extreme value attention network for long-term air quality
  prediction.
\newblock In \emph{2020 IEEE International Conference on Big Data (Big Data)},
  pages 4545--4552. IEEE, 2020.

\bibitem[Zhang et~al.(2021)Zhang, Li, Carlo, Manda, Hamshaw, Dascalu, Harris,
  and Wu]{Yifan2021}
Yifan Zhang, Jiahao Li, Ablan Carlo, Alex~K Manda, Scott Hamshaw, Sergiu~M.
  Dascalu, Frederick~C. Harris, and Rui Wu.
\newblock Data regression framework for time series data with extreme events.
\newblock In \emph{2021 IEEE International Conference on Big Data (Big Data)},
  pages 5327--5336, 2021.
\newblock \doi{10.1109/BigData52589.2021.9671387}.

\bibitem[Li et~al.(2023{\natexlab{a}})Li, Xu, and
  Anastasiu]{li2022extremeadaptive}
Yanhong Li, Jack Xu, and David~C. Anastasiu.
\newblock An extreme-adaptive time series prediction model based on
  probability-enhanced lstm neural networks.
\newblock In \emph{Proceedings of the AAAI Conference on Artificial
  Intelligence}, 2023{\natexlab{a}}.

\bibitem[Box and Pierce(1970)]{box1970distribution}
George~EP Box and David~A Pierce.
\newblock Distribution of residual autocorrelations in
  autoregressive-integrated moving average time series models.
\newblock \emph{Journal of the American statistical Association}, 65\penalty0
  (332):\penalty0 1509--1526, 1970.

\bibitem[Nielsen(2019)]{nielsen2019practical}
Aileen Nielsen.
\newblock \emph{Practical time series analysis: Prediction with statistics and
  machine learning}.
\newblock O'Reilly Media, 2019.

\bibitem[Box and Jenkins(1976)]{boxjen76}
George.E.P. Box and Gwilym~M. Jenkins.
\newblock \emph{Time Series Analysis: Forecasting and Control}.
\newblock Holden-Day, {}, 1976.

\bibitem[Shortridge et~al.(2016)Shortridge, Guikema, and
  Zaitchik]{shortridge2016machine}
Julie~E Shortridge, Seth~D Guikema, and Benjamin~F Zaitchik.
\newblock Machine learning methods for empirical streamflow simulation: a
  comparison of model accuracy, interpretability, and uncertainty in seasonal
  watersheds.
\newblock \emph{Hydrology and Earth System Sciences}, 20\penalty0 (7):\penalty0
  2611--2628, 2016.

\bibitem[Papacharalampous and Tyralis(2022)]{papacharalampous2022review}
Georgia Papacharalampous and Hristos Tyralis.
\newblock A review of machine learning concepts and methods for addressing
  challenges in probabilistic hydrological post-processing and forecasting.
\newblock \emph{Frontiers in Water}, 4:\penalty0 961954, 2022.

\bibitem[Cheng et~al.(2020)Cheng, Fang, Kinouchi, Navon, and
  Pain]{cheng2020long}
M~Cheng, F~Fang, T~Kinouchi, IM~Navon, and CC~Pain.
\newblock Long lead-time daily and monthly streamflow forecasting using machine
  learning methods.
\newblock \emph{Journal of Hydrology}, 590:\penalty0 125376, 2020.

\bibitem[Lai et~al.(2018)Lai, Chang, Yang, and Liu]{lai2018modeling}
Guokun Lai, Wei-Cheng Chang, Yiming Yang, and Hanxiao Liu.
\newblock Modeling long-and short-term temporal patterns with deep neural
  networks.
\newblock In \emph{The 41st international ACM SIGIR conference on research \&
  development in information retrieval}, pages 95--104, 2018.

\bibitem[Siami-Namini et~al.(2018)Siami-Namini, Tavakoli, and
  Namin]{siami2018comparison}
Sima Siami-Namini, Neda Tavakoli, and Akbar~Siami Namin.
\newblock A comparison of arima and lstm in forecasting time series.
\newblock In \emph{2018 17th IEEE international conference on machine learning
  and applications (ICMLA)}, pages 1394--1401. IEEE, 2018.

\bibitem[Oreshkin et~al.(2019)Oreshkin, Carpov, Chapados, and
  Bengio]{oreshkin2019n}
Boris~N Oreshkin, Dmitri Carpov, Nicolas Chapados, and Yoshua Bengio.
\newblock N-beats: Neural basis expansion analysis for interpretable time
  series forecasting.
\newblock \emph{arXiv preprint arXiv:1905.10437}, 2019.

\bibitem[Wu et~al.(2020)Wu, Pan, Long, Jiang, Chang, and
  Zhang]{wu2020connecting}
Zonghan Wu, Shirui Pan, Guodong Long, Jing Jiang, Xiaojun Chang, and Chengqi
  Zhang.
\newblock Connecting the dots: Multivariate time series forecasting with graph
  neural networks.
\newblock In \emph{Proceedings of the 26th ACM SIGKDD international conference
  on knowledge discovery \& data mining}, pages 753--763, 2020.

\bibitem[Cao et~al.(2020)Cao, Wang, Duan, Zhang, Zhu, Huang, Tong, Xu, Bai,
  Tong, et~al.]{cao2020spectral}
Defu Cao, Yujing Wang, Juanyong Duan, Ce~Zhang, Xia Zhu, Congrui Huang, Yunhai
  Tong, Bixiong Xu, Jing Bai, Jie Tong, et~al.
\newblock Spectral temporal graph neural network for multivariate time-series
  forecasting.
\newblock \emph{Advances in neural information processing systems},
  33:\penalty0 17766--17778, 2020.

\bibitem[Li et~al.(2019)Li, Jin, Xuan, Zhou, Chen, Wang, and
  Yan]{li2019enhancing}
Shiyang Li, Xiaoyong Jin, Yao Xuan, Xiyou Zhou, Wenhu Chen, Yu-Xiang Wang, and
  Xifeng Yan.
\newblock Enhancing the locality and breaking the memory bottleneck of
  transformer on time series forecasting.
\newblock \emph{Advances in neural information processing systems}, 32, 2019.

\bibitem[Qin et~al.(2017)Qin, Song, Chen, Cheng, Jiang, and
  Cottrell]{qin2017dual}
Yao Qin, Dongjin Song, Haifeng Chen, Wei Cheng, Guofei Jiang, and Garrison
  Cottrell.
\newblock A dual-stage attention-based recurrent neural network for time series
  prediction.
\newblock \emph{arXiv preprint arXiv:1704.02971}, 2017.

\bibitem[Zhou et~al.(2021)Zhou, Zhang, Peng, Zhang, Li, Xiong, and
  Zhang]{zhou2021informer}
Haoyi Zhou, Shanghang Zhang, Jieqi Peng, Shuai Zhang, Jianxin Li, Hui Xiong,
  and Wancai Zhang.
\newblock Informer: Beyond efficient transformer for long sequence time-series
  forecasting.
\newblock In \emph{Proceedings of the AAAI conference on artificial
  intelligence}, volume~35, pages 11106--11115, 2021.

\bibitem[Zhou et~al.(2022)Zhou, Ma, Wen, Wang, Sun, and Jin]{zhou2022fedformer}
Tian Zhou, Ziqing Ma, Qingsong Wen, Xue Wang, Liang Sun, and Rong Jin.
\newblock Fedformer: Frequency enhanced decomposed transformer for long-term
  series forecasting.
\newblock In \emph{International Conference on Machine Learning}, pages
  27268--27286. PMLR, 2022.

\bibitem[Kitaev et~al.(2020)Kitaev, Kaiser, and Levskaya]{kitaev2020reformer}
Nikita Kitaev, {\L}ukasz Kaiser, and Anselm Levskaya.
\newblock Reformer: The efficient transformer.
\newblock \emph{arXiv preprint arXiv:2001.04451}, 2020.

\bibitem[Zeng et~al.(2022)Zeng, Chen, Zhang, and Xu]{zeng2022transformers}
Ailing Zeng, Muxi Chen, Lei Zhang, and Qiang Xu.
\newblock Are transformers effective for time series forecasting?
\newblock \emph{arXiv preprint arXiv:2205.13504}, 2022.

\bibitem[Das et~al.(2023)Das, Kong, Leach, Sen, and Yu]{das2023long}
Abhimanyu Das, Weihao Kong, Andrew Leach, Rajat Sen, and Rose Yu.
\newblock Long-term forecasting with tide: Time-series dense encoder.
\newblock \emph{arXiv preprint arXiv:2304.08424}, 2023.

\bibitem[Fortuin et~al.(2018)Fortuin, H{\"u}ser, Locatello, Strathmann, and
  R{\"a}tsch]{fortuin2018som}
Vincent Fortuin, Matthias H{\"u}ser, Francesco Locatello, Heiko Strathmann, and
  Gunnar R{\"a}tsch.
\newblock Som-vae: Interpretable discrete representation learning on time
  series.
\newblock \emph{arXiv preprint arXiv:1806.02199}, 2018.

\bibitem[Lei et~al.(2017)Lei, Yi, Vaculin, Wu, and Dhillon]{lei2017similarity}
Qi~Lei, Jinfeng Yi, Roman Vaculin, Lingfei Wu, and Inderjit~S Dhillon.
\newblock Similarity preserving representation learning for time series
  clustering.
\newblock \emph{arXiv preprint arXiv:1702.03584}, 2017.

\bibitem[Tonekaboni et~al.(2022)Tonekaboni, Li, Arik, Goldenberg, and
  Pfister]{tonekaboni2022decoupling}
Sana Tonekaboni, Chun-Liang Li, Sercan~O Arik, Anna Goldenberg, and Tomas
  Pfister.
\newblock Decoupling local and global representations of time series.
\newblock In \emph{International Conference on Artificial Intelligence and
  Statistics}, pages 8700--8714. PMLR, 2022.

\bibitem[Ma et~al.(2021)Ma, Zhang, Li, and Lu]{ma2021unsupervised}
Haojie Ma, Zhijie Zhang, Wenzhong Li, and Sanglu Lu.
\newblock Unsupervised human activity representation learning with multi-task
  deep clustering.
\newblock \emph{Proceedings of the ACM on Interactive, Mobile, Wearable and
  Ubiquitous Technologies}, 5\penalty0 (1):\penalty0 1--25, 2021.

\bibitem[Ma et~al.(2018)Ma, Zhao, Yi, Chen, Hong, and Chi]{ma2018modeling}
Jiaqi Ma, Zhe Zhao, Xinyang Yi, Jilin Chen, Lichan Hong, and Ed~H Chi.
\newblock Modeling task relationships in multi-task learning with multi-gate
  mixture-of-experts.
\newblock In \emph{Proceedings of the 24th ACM SIGKDD international conference
  on knowledge discovery \& data mining}, pages 1930--1939, 2018.

\bibitem[Wang et~al.(2018)Wang, Qiu, and Li]{w10070853}
Zhi-Yu Wang, Jun Qiu, and Fang-Fang Li.
\newblock Hybrid models combining emd/eemd and arima for long-term streamflow
  forecasting.
\newblock \emph{Water}, 10\penalty0 (7), 2018.
\newblock ISSN 2073-4441.
\newblock \doi{10.3390/w10070853}.

\bibitem[Sen et~al.(2019)Sen, Yu, and Dhillon]{sen2019think}
Rajat Sen, Hsiang-Fu Yu, and Inderjit~S Dhillon.
\newblock Think globally, act locally: A deep neural network approach to
  high-dimensional time series forecasting.
\newblock \emph{Advances in neural information processing systems}, 32, 2019.

\bibitem[Box et~al.(2015)Box, Jenkins, Reinsel, and Ljung]{box2015time}
George~EP Box, Gwilym~M Jenkins, Gregory~C Reinsel, and Greta~M Ljung.
\newblock \emph{Time series analysis: forecasting and control}.
\newblock John Wiley \& Sons, 2015.

\bibitem[Salinas et~al.(2020)Salinas, Flunkert, Gasthaus, and
  Januschowski]{salinas2020deepar}
David Salinas, Valentin Flunkert, Jan Gasthaus, and Tim Januschowski.
\newblock Deepar: Probabilistic forecasting with autoregressive recurrent
  networks.
\newblock \emph{International Journal of Forecasting}, 36\penalty0
  (3):\penalty0 1181--1191, 2020.

\bibitem[Wu et~al.(2021)Wu, Xu, Wang, and Long]{wu2021autoformer}
Haixu Wu, Jiehui Xu, Jianmin Wang, and Mingsheng Long.
\newblock Autoformer: Decomposition transformers with auto-correlation for
  long-term series forecasting.
\newblock \emph{Advances in Neural Information Processing Systems},
  34:\penalty0 22419--22430, 2021.

\bibitem[Singh et~al.(2022)Singh, Ranjan, and Tiwari]{singh2022credit}
Amit Singh, Ranjeet~Kumar Ranjan, and Abhishek Tiwari.
\newblock Credit card fraud detection under extreme imbalanced data: a
  comparative study of data-level algorithms.
\newblock \emph{Journal of Experimental \& Theoretical Artificial
  Intelligence}, 34\penalty0 (4):\penalty0 571--598, 2022.

\bibitem[An and Cho(2015)]{an2015variational}
Jinwon An and Sungzoon Cho.
\newblock Variational autoencoder based anomaly detection using reconstruction
  probability.
\newblock \emph{Special lecture on IE}, 2\penalty0 (1):\penalty0 1--18, 2015.

\bibitem[Li et~al.(2023{\natexlab{b}})Li, Xu, and Anastasiu]{li2023BigData}
Yanhong Li, Jack Xu, and David~C. Anastasiu.
\newblock Seed: An effective model for highly-skewed streamflow time series
  data forecasting.
\newblock In \emph{2023 IEEE International Conference on Big Data (Big Data)},
  IEEE BigData 2023, Los Alamitos, CA, USA, Dec 2023{\natexlab{b}}. IEEE
  Computer Society.

\bibitem[Day(1969{\natexlab{a}})]{day1969}
N.~E. Day.
\newblock {Estimating the components of a mixture of normal distributions}.
\newblock \emph{Biometrika}, 56\penalty0 (3):\penalty0 463--474, 12
  1969{\natexlab{a}}.
\newblock ISSN 0006-3444.
\newblock \doi{10.1093/biomet/56.3.463}.

\bibitem[McKight and Najab(2010)]{mckight2010kruskal}
Patrick~E McKight and Julius Najab.
\newblock Kruskal-wallis test.
\newblock \emph{The corsini encyclopedia of psychology}, pages 1--1, 2010.

\bibitem[Ostertagova et~al.(2014)Ostertagova, Ostertag, and
  Kov{\'a}{\v{c}}]{ostertagova2014methodology}
Eva Ostertagova, Oskar Ostertag, and Jozef Kov{\'a}{\v{c}}.
\newblock Methodology and application of the kruskal-wallis test.
\newblock In \emph{Applied mechanics and materials}, volume 611, pages
  115--120. Trans Tech Publ, 2014.

\bibitem[Breslow(1970)]{breslow1970generalized}
Norman Breslow.
\newblock A generalized kruskal-wallis test for comparing k samples subject to
  unequal patterns of censorship.
\newblock \emph{Biometrika}, 57\penalty0 (3):\penalty0 579--594, 1970.

\bibitem[Le et~al.(2021)Le, Chen, Hang, and Hu]{le2021attention}
Yan Le, Changwei Chen, Ting Hang, and Youchuan Hu.
\newblock A stream prediction model based on attention-lstm.
\newblock \emph{Earth Science Informatics}, 14:\penalty0 1--11, 06 2021.
\newblock \doi{10.1007/s12145-021-00571-z}.

\bibitem[Makridakis and Hibon(2000)]{MAKRIDAKIS2000451}
Spyros Makridakis and Michèle Hibon.
\newblock The m3-competition: results, conclusions and implications.
\newblock \emph{International Journal of Forecasting}, 16\penalty0
  (4):\penalty0 451--476, 2000.
\newblock ISSN 0169-2070.
\newblock \doi{https://doi.org/10.1016/S0169-2070(00)00057-1}.
\newblock The M3- Competition.

\bibitem[Makridakis et~al.(2018)Makridakis, Spiliotis, and
  Assimakopoulos]{makridakis2018m4}
Spyros Makridakis, Evangelos Spiliotis, and Vassilios Assimakopoulos.
\newblock The m4 competition: Results, findings, conclusion and way forward.
\newblock \emph{International Journal of Forecasting}, 34\penalty0
  (4):\penalty0 802--808, 2018.

\bibitem[Athanasopoulos et~al.(2011)Athanasopoulos, Hyndman, Song, and
  Wu]{athanasopoulos2011tourism}
George Athanasopoulos, Rob~J Hyndman, Haiyan Song, and Doris~C Wu.
\newblock The tourism forecasting competition.
\newblock \emph{International Journal of Forecasting}, 27\penalty0
  (3):\penalty0 822--844, 2011.

\bibitem[Nayak et~al.(2004)Nayak, Sudheer, Rangan, and Ramasastri]{NAYAK200452}
P.C Nayak, K.P Sudheer, D.M Rangan, and K.S Ramasastri.
\newblock A neuro-fuzzy computing technique for modeling hydrological time
  series.
\newblock \emph{Journal of Hydrology}, 291\penalty0 (1):\penalty0 52--66, 2004.
\newblock ISSN 0022-1694.
\newblock \doi{https://doi.org/10.1016/j.jhydrol.2003.12.010}.

\bibitem[Tabbussum and Dar(2021)]{tabbussum2021comparison}
Ruhhee Tabbussum and Abdul~Qayoom Dar.
\newblock Comparison of fuzzy inference algorithms for stream flow prediction.
\newblock \emph{Neural Computing and Applications}, 33:\penalty0 1643--1653,
  2021.

\bibitem[Nhu et~al.(2020)Nhu, Shahabi, Nohani, Shirzadi, Al-Ansari, Bahrami,
  Miraki, Geertsema, and Nguyen]{RFbased}
Viet-Ha Nhu, Himan Shahabi, Ebrahim Nohani, Ataollah Shirzadi, Nadhir
  Al-Ansari, Sepideh Bahrami, Shaghayegh Miraki, Marten Geertsema, and Hoang
  Nguyen.
\newblock Daily water level prediction of zrebar lake (iran): A comparison
  between m5p, random forest, random tree and reduced error pruning trees
  algorithms.
\newblock \emph{ISPRS International Journal of Geo-Information}, 9\penalty0
  (8), 2020.
\newblock ISSN 2220-9964.
\newblock \doi{10.3390/ijgi9080479}.

\bibitem[Day(1969{\natexlab{b}})]{day1969-a}
N.~E. Day.
\newblock {Estimating the components of a mixture of normal distributions}.
\newblock \emph{Biometrika}, 56\penalty0 (3):\penalty0 463--474, 12
  1969{\natexlab{b}}.
\newblock ISSN 0006-3444.
\newblock \doi{10.1093/biomet/56.3.463}.

\end{thebibliography}

\end{document}